\documentclass[11pt]{scaleai-paper}

\usepackage{amsmath,amsfonts,amssymb,booktabs,multirow,subcaption,float,array}
\usepackage[edges]{forest}
\usepackage{tabularx,tabulary}
\usepackage[square,numbers,sort&compress]{natbib}
\usepackage{xspace,url,xcolor}
\usepackage[colorlinks=true,linkcolor=scaleLink,citecolor=scaleLink,urlcolor=scaleLink]{hyperref}
\usepackage[capitalise,nameinlink]{cleveref}
\usepackage{placeins}
\usepackage{needspace}

\usepackage{amsmath,amsfonts,bm}

\def\eqref#1{equation~\ref{#1}}

\def\1{\bm{1}}

\DeclareMathAlphabet{\mathsfit}{\encodingdefault}{\sfdefault}{m}{sl}
\SetMathAlphabet{\mathsfit}{bold}{\encodingdefault}{\sfdefault}{bx}{n}

\newcommand{\forge}{\textbf{RSI-Forge}\xspace}
\newcommand{\nSourceYear}{2026}

\newif\ifdraftnums
\draftnumsfalse
\newcommand{\provisional}[1]{\ifdraftnums{\color{red}#1}\else#1\fi}

\usepackage{longtable}

\newcolumntype{Y}{>{\RaggedRight\arraybackslash}X}

\contact{\texttt{\{renxiong.wang,darvin.yi,tong.zhao,yunzhong.he\}@scale.com}}

\title{\fontsize{23}{28}\selectfont RSI-Forge: From Research Papers to Environments for Recursive Self-Improvement}
\author[1]{Renxiong Wang}
\author[1]{Darvin Yi}
\author[1]{Abril Herrlein}
\author[1]{Anas Mahmoud}
\author[1]{Advait Gosai}
\author[1]{Lisiman Hua}
\author[1]{MohammadHossein Rezaei}
\author[1]{Xingang Guo}
\author[1]{Anisha Gunjal}
\author[1]{Utkarsh Tyagi}
\author[1]{David J. Lee}
\author[1]{Minglai Yang}
\author[1]{Haris Riaz}
\author[1,2]{Chenguang Wang}
\author[3]{Huaxiu Yao}
\author[1]{Daniel Yue Zhang}
\author[1]{Aakash Sabharwal}
\author[1]{Tong Zhao}
\author[1]{Yunzhong He}
\affil[1]{Scale AI}
\affil[2]{University of California, Santa Cruz}
\affil[3]{University of North Carolina at Chapel Hill}

\newcommand{\nAblJudgeGemCor}{\provisional{4.88}}
\newcommand{\nAblJudgeGemCorN}{\provisional{180}}
\newcommand{\nAblJudgeGemSol}{\provisional{4.74}}
\newcommand{\nAblJudgeGemSolN}{\provisional{30}}
\newcommand{\nAblJudgeGptCor}{\provisional{4.40}}
\newcommand{\nAblJudgeGptCorN}{\provisional{180}}
\newcommand{\nAblJudgeGptSol}{\provisional{4.27}}
\newcommand{\nAblJudgeGptSolN}{\provisional{30}}
\newcommand{\nAblJudgeOpusCor}{\provisional{4.37}}
\newcommand{\nAblJudgeOpusCorN}{\provisional{180}}
\newcommand{\nAblJudgeOpusSol}{\provisional{4.25}}
\newcommand{\nAblJudgeOpusSolN}{\provisional{30}}
\newcommand{\nAblSolN}{\provisional{30}}
\newcommand{\nBeatDeep}{54}
\newcommand{\nBeatEnvs}{\provisional{68}}
\newcommand{\nBeatGpt}{65}
\newcommand{\nBeatHaiku}{24}
\newcommand{\nBeatOpus}{63}
\newcommand{\nBeatRetuneHaiku}{77}
\newcommand{\nBeatRetuneOpus}{34}
\newcommand{\nCandidates}{\provisional{2,117}}
\newcommand{\nCodeSample}{\provisional{24}}
\newcommand{\nCoderCats}{9}
\newcommand{\nCoderExplHaiku}{13}
\newcommand{\nCoderExplOpus}{93}
\newcommand{\nCoderKappa}{\provisional{0.73}}
\newcommand{\nCoderPerModel}{\provisional{30}}
\newcommand{\nCoderShortHaiku}{63}
\newcommand{\nCoderShortOpus}{0}
\newcommand{\nCommonEnvs}{\provisional{120}}
\newcommand{\nDelivered}{\provisional{180}}
\newcommand{\nErrBarGens}{\provisional{139}}
\newcommand{\nErrBarHaiku}{\provisional{73}}
\newcommand{\nErrBarOpus}{\provisional{44}}
\newcommand{\nFamMapped}{\provisional{62}}
\newcommand{\nFieldMax}{\provisional{14}}
\newcommand{\nFieldMin}{\provisional{6}}
\newcommand{\nFields}{\provisional{18}}
\newcommand{\nGapfaithful}{\provisional{0.50}}
\newcommand{\nGapnotsaturable}{\provisional{0.89}}
\newcommand{\nGapungameable}{\provisional{0.58}}
\newcommand{\nHandBuiltDims}{\provisional{8}}
\newcommand{\nHandBuiltGap}{\provisional{+0.46}}
\newcommand{\nHandBuiltMean}{\provisional{4.11}}
\newcommand{\nHandBuiltN}{\provisional{34}}
\newcommand{\nHandBuiltOurs}{\provisional{4.58}}
\newcommand{\nHarnessDiff}{\provisional{0.13}}
\newcommand{\nHarnessEnvs}{\provisional{180}}
\newcommand{\nHarnessMeanAgy}{\provisional{4.82}}
\newcommand{\nHarnessMeanCc}{\provisional{4.88}}
\newcommand{\nHarnessPairs}{\provisional{1,620}}
\newcommand{\nHarnessWithin}{\provisional{99}}
\newcommand{\nHdiscriminating}{\provisional{4.74}}
\newcommand{\nHheadroom}{\provisional{4.81}}
\newcommand{\nHnotsaturable}{\provisional{3.09}}
\newcommand{\nHumanEnvs}{\provisional{90}}
\newcommand{\nHumanMean}{\provisional{4.24}}
\newcommand{\nJudgeDims}{\provisional{9}}
\newcommand{\nPassBuild}{\provisional{198}}
\newcommand{\nPassDesign}{\provisional{199}}
\newcommand{\nPassReproduce}{\provisional{189}}
\newcommand{\nPassSmoke}{180}
\newcommand{\nPnotsaturable}{\provisional{3.98}}
\newcommand{\nReachDesign}{\provisional{213}}
\newcommand{\nRepairWorst}{\provisional{3}}
\newcommand{\nRoundsN}{\provisional{3}}
\newcommand{\nScreened}{\provisional{10,800}}
\newcommand{\nWinMechs}{\provisional{9}}
\newcommand{\nWinNeither}{0}
\newcommand{\nWinStructural}{95}
\newcommand{\nWinTuningOnly}{5}
\newcommand{\nFamImprove}{\provisional{21}}
\newcommand{\nFamExplore}{\provisional{27}}
\newcommand{\nFamShort}{\provisional{14}}
\newcommand{\nRoundBudgetMin}{\provisional{60}}
\newcommand{\nPanelGens}{360}
\newcommand{\nPanelMeanThree}{4.54}
\newcommand{\nPanelMeanTwo}{4.39}
\newcommand{\nRepairAnyPct}{79}
\newcommand{\nRepairDesignPct}{53}
\newcommand{\nRepairBuildPct}{45}
\newcommand{\nRepairReproPct}{21}
\newcommand{\nRepairSmokePct}{20}
\newcommand{\nLaterEnvPct}{84}
\newcommand{\nCoderExplOpusOne}{97}
\newcommand{\nCoderExplHaikuOne}{30}
\newcommand{\nCoderShortOpusOne}{3}
\newcommand{\nCoderShortHaikuOne}{80}
\newcommand{\nRefLost}{0.76}
\newcommand{\nRefWon}{0.46}
\newcommand{\nCleanA}{0.797}
\newcommand{\nCleanB}{0.761}
\newcommand{\nCleanC}{0.641}
\newcommand{\nCleanD}{0.332}
\newcommand{\nCostTotalM}{170}
\newcommand{\nCostAuthorPct}{82}
\newcommand{\nCostReviewerPct}{14}
\newcommand{\nCostReproducerPct}{3}
\newcommand{\nFlaggedChains}{22}
\newcommand{\nPanelChains}{480}
\newcommand{\nCleanShift}{0.006}
\newcommand{\nWinsRead}{4}

\begingroup
\renewcommand{\provisional}[1]{#1}
\xdef\nTotalDeliveredRaw{\the\numexpr\nDelivered+\nAblSolN\relax}
\endgroup
\newcommand{\nTotalDelivered}{\provisional{\nTotalDeliveredRaw}}

\usepackage{enumitem}
\newcommand{\gptfivesol}{GPT-5.6-sol\xspace}
\newcommand{\opusfive}{Claude Opus 5\xspace}
\newcommand{\haikufive}{Claude Haiku 4.5\xspace}
\newcommand{\gptfive}{GPT-5.6-sol\xspace}
\newcommand{\deepseek}{DeepSeek-V4-Flash\xspace}
\newcommand{\sonnetfive}{Claude Sonnet 5\xspace}
\newcommand{\geminiflash}{Gemini 3.7 Flash\xspace}
\begin{document}
\maketitle

\begin{abstract}
Environments are the foundation of recursive self-improvement: they provide the problems agents work on and the feedback used to evaluate progress. Yet constructing challenging research environments with reliable evaluation still depends on domain experts, limiting their scale and disciplinary coverage. We introduce \forge{}, a multi-agent pipeline that turns published papers into executable environments for self-improvement. Three agents coordinate construction, reproduction, and review to produce tasks with automated evaluators; each paper's method is independently reimplemented to establish a baseline score.
We present \nTotalDelivered{} environments across \nFields{} fields, including \nHumanEnvs{} reviewed by independent human domain experts.
Both experts and agent judges give high ratings to the potential for improving the provided starting solutions and the evaluators' ability to distinguish solution quality, whereas experts are more critical of shortcut resistance, faithfulness to the source paper, and whether a single idea can exhaust a task.
To validate their use for repeated improvement, we evaluate four models over \nRoundsN{} successive attempts on \nCommonEnvs{} environments, with each attempt inheriting prior code and notes while model weights remain fixed.
At least one model improves after the first attempt in \nLaterEnvPct{}\% of environments.
Models also outperform the reproduced paper methods in \nBeatEnvs{} of the \nCommonEnvs{} environments, demonstrating room for gains beyond these baselines.
Transcript analysis identifies work beyond parameter tuning in \nWinStructural{}\% of these successful attempts.
Analysis of the resulting trajectories shows that models scoring lower on these tasks explore less, more often accept gains smaller than the reported standard error, and rely more heavily on tuning to the development set.
\forge{} provides a scalable approach to constructing research environments for training and evaluating self-improving agents.
\end{abstract}

\section{Introduction}\label{sec:intro}

Recursive self-improvement (RSI) is a process in which an agent improves its own capabilities and uses those improvements to guide further changes \citep{rsisurvey2026}. Such agents could accumulate useful strategies and tackle increasingly difficult problems with less human intervention. Developing this capability requires environments that provide challenging problems and reliable feedback on attempted improvements. These environments supply experience for training and a basis for evaluation, making their availability and diversity central to the development of RSI at scale.

Constructing suitable RSI environments is particularly difficult for research tasks. They must pose meaningful problems, allow continued improvement, and evaluate solutions automatically while distinguishing genuine gains from measurement noise and scoring shortcuts. Formulating such tasks requires domain knowledge, and implementing reliable evaluators adds a separate engineering challenge. Existing benchmarks draw on repositories and competitions with established tests or graders \citep{jimenez2024swebench, chan2025mlebench}, or rely on tasks and evaluation rubrics written by experts \citep{starace2025paperbench, wijk2024rebench}. These approaches provide valuable environments, but their scale and domain coverage remain tied to available evaluation resources and expert effort.
This raises a central question: \emph{how can we automatically construct RSI environments at scale?}

\begin{figure}[t]\centering
\includegraphics[width=1\textwidth]{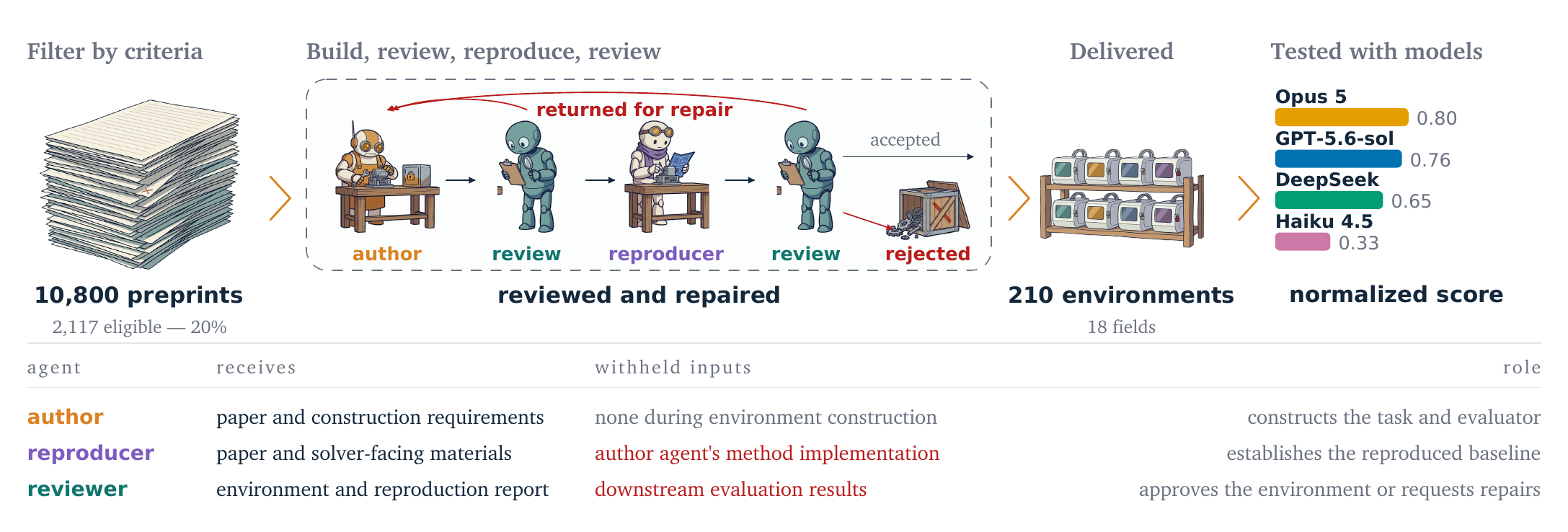}
\caption{Overview of \forge{} and the information available to each agent. Failed checks lead to repair or rejection before an environment is accepted. The output count includes all \nTotalDelivered{} environments. The performance bars report median scores of final solutions after \nRoundsN{} successive attempts on \nCommonEnvs{} environments sampled from the main construction run. Scores are normalized so that the starting solution is 0 and the target score is 1 (\cref{eq:band}).
}
\label{fig:teaser}
\end{figure}

We introduce \forge{}, a multi-agent pipeline that constructs executable RSI environments from published papers (\cref{fig:teaser}).
Papers provide concrete research questions, methods, and measured results that can guide environment design. An author agent develops the task, starting solution, and evaluator. A reproducer agent independently implements the paper's method using only the paper and solver-facing materials, establishing the reproduced baseline. A reviewer agent checks that the task can be understood from its instructions and that the evaluator measures the stated objective, requesting repairs when needed. The agents have distinct responsibilities and access to information, allowing reproduction and review to assess the environment separately from the author agent's construction.

Using this pipeline, we construct \nTotalDelivered{} environments across \nFields{} fields.
To assess their quality, three agent judges evaluate the environments using a common rubric, and independent human domain experts review \nHumanEnvs{} environments within their fields of expertise. The rubric assesses whether tasks pose meaningful research problems, offer opportunities for improvement, and provide reliable evaluation. Experts assign their highest ratings to the potential for improving the starting solutions and the evaluators' ability to distinguish solution quality. They are more critical than agent judges of shortcut resistance, fidelity to the source paper, and whether a single idea can exhaust a task (\cref{fig:expert}).

To assess whether these environments support repeated improvement, we evaluate four solver models over \nRoundsN{} successive attempts on \nCommonEnvs{} of the environments. Each attempt inherits prior code and notes while model weights remain fixed. At least one model improves after the first attempt in \nLaterEnvPct{}\% of environments, and the reproduced baseline is surpassed in \nBeatEnvs{} of the \nCommonEnvs{} environments.
Transcript analysis identifies work beyond parameter tuning in \nWinStructural{}\% of these successful attempts.
Analysis of the resulting trajectories further distinguishes how models explore alternatives, assess uncertain gains, and select solutions. Together, these results show that \forge{} can automatically construct research environments that support measurable improvement over successive attempts. Our contributions are:

\begin{enumerate}[leftmargin=0.4cm,nosep]
\item \textbf{An automated RSI environment construction pipeline.} We introduce RSI-Forge, which turns published papers into executable environments.
\item \textbf{A corpus spanning diverse research fields.} We construct \nTotalDelivered{} environments across \nFields{} fields, including \nHumanEnvs{} independently reviewed by human domain experts.
\item \textbf{Validation through repeated improvement.} We empirically demonstrate gains over successive attempts and beyond reproduced baselines, with trajectory analysis revealing differences in how models explore alternatives and assess changes.
\end{enumerate}

\section{Related Work}\label{sec:related}

\textbf{Self-improving agents.}
Prior work studies improvement at several levels. Self-Refine iteratively revises model outputs using feedback, while Reflexion retains reflections from earlier attempts to inform subsequent decisions \citep{madaan2023selfrefine, shinn2023reflexion}. Other approaches optimize prompts, workflows, agent designs, harnesses, or source code \citep{promptbreeder2023, aflow2024, hu2024adas, zelikman2024stop, dgm2025, godelagent2024, rhi2026}. These approaches investigate how agents improve their solutions and procedures. Our work addresses the construction of environments that supply challenging problems and measurable feedback for such improvement.

\textbf{Automated task and environment generation.}
Research on learning tasks spans curriculum learning, adaptive environment design, and autonomous task selection \citep{bengio2009curriculum, dennis2020paired, jiang2021plr, wang2023voyager}. Absolute Zero and R-Zero generate tasks as part of a learning process that couples task proposal with solving \citep{absolutezero2025, rzero2025}. AgentSynth synthesizes tasks and trajectories for computer-use agents \citep{agentsynth2025}, while EnvScaler and EnvFactory construct executable environments for tool interaction \citep{envscaler2026, envfactory2026}. SPADE further couples environment generation with learning: a model designs executable environments and learns by acting within them \citep{spade2026}. For terminal agents, TermiGen synthesizes verifiable environments through iterative multi-agent refinement \citep{termigen2026}, while \citet{envevolution2026} progressively increase environment difficulty to sustain learning signals. \forge{} shares the goal of expanding environment supply through automation, focusing on research problems drawn from published papers across diverse fields.

\textbf{Research and engineering benchmarks.}
Existing benchmarks obtain tasks and evaluation procedures from several sources. SWE-bench uses repository issues and tests, while MLE-bench adapts machine-learning competitions \citep{jimenez2024swebench, chan2025mlebench}. Related benchmarks assess machine-learning experimentation and computational reproducibility \citep{huang2024mlagentbench, siegel2024corebench}. PaperBench evaluates paper replication using rubrics developed with the source papers' authors \citep{starace2025paperbench}; RE-Bench provides expert-designed research engineering environments \citep{wijk2024rebench}; and Terminal-Bench provides curated terminal tasks with human-written solutions and verification tests \citep{termbench2026}. RSI-Exam focuses specifically on repeated improvement in executable research tasks \citep{rsiexam2026}. Closer to automated construction from papers, \citet{theiler2026} translate published methods into a shared benchmarking framework for machine health intelligence. \forge{} constructs research tasks and evaluators across fields, with independently reproduced baselines for measuring further improvements.

\section{RSI-Forge}
\label{sec:pipeline}

RSI-Forge draws research problems and reference methods from published papers. It converts each selected problem into an executable environment and independently reimplements the paper's method there to establish a measured baseline. A reviewer agent checks that the instructions clearly describe the task and the evaluator measures the intended objective, requesting repairs when needed.

\subsection{Environment Format}\label{sec:env}\label{sec:data}

We start by defining an RSI environment as an executable research task with the resources needed to develop and evaluate solutions. Each environment specifies a research objective, the required solution interface, and a compute budget. It provides a runnable starting solution, code for generating task data, and an automated evaluator. We refer to the agent attempting the task as the solver. The solver can modify or replace the starting solution and test its changes on a development set. A separate held-out test set is reserved for grading submitted solutions.

Each environment includes a reproduced baseline: the source paper's method independently reimplemented and evaluated under the environment's task definition and resource limits. Its measured score provides a reference for assessing improvements beyond that method. The source paper, reproduced implementation, and test data are not supplied as solver inputs.

Each environment also defines a target score based on a task-specific performance bound or an idealized solution, such as one with access to information unavailable to the solver. The target serves as a normalization reference and need not be attainable with the information and resources available to the solver. To compare performance across environments, we normalize scores so that the starting solution corresponds to 0 and the target to 1.

\subsection{Multi-Agent Construction}

Construction begins with papers whose research problems can be evaluated automatically and whose methods fit the available compute budget. For each paper, an author agent adapts the research problem into an executable task, specifying the solver's objective, permitted changes, and resource limits. It builds the starting solution, code for generating task data, and evaluator, and prepares the development and held-out test sets. The author agent also implements the paper's method for preliminary checks and measures the starting-solution and target scores.

The proposed environment then undergoes automated checks and review. A reviewer agent, with a context separate from the author agent's, examines the instructions, implementation, and check results to assess whether the task can be understood from the supplied materials and whether the evaluator measures the intended research objective. Initial checks verify that the components run and that the evaluator executes the code submitted through the required interface. Scripted tests also check whether performance differences exceed measured variation across repeated evaluations. The author agent specifies how the test cases differ from the development cases---for example, by reserving different simulation settings for testing---and the reviewer agent checks this distinction.

Once these checks pass, a reproducer agent receives the paper together with the task instructions, development data, and evaluator available to a solver. The author agent's implementation of the method, construction history, the test data, and the target score are withheld. The reproducer agent independently implements the paper's method using the same submission interface and compute budget as solver solutions. Its implementation is evaluated on the held-out test set to establish the reproduced baseline score. The reproduction must outperform the starting solution for the environment to pass this stage, providing direct evidence that improvement over the starting solution is possible.

The reviewer agent then examines the reproduction report, and a final end-to-end test checks that the environment can start, accept a solver's submission, and grade it. Acceptance requires successful automated checks and approval from the reviewer agent at every stage. A failed check or unresolved review finding returns the environment to the author agent for repair, and environments that still fail after the permitted repair attempts are rejected.

\subsection{Scoring and Verification}\label{sec:band}

\textbf{Scoring solver submissions.}
Each environment measures its research objective using a scoring program that is deterministic for fixed evaluation inputs. To enable performance comparison across environments, we normalize a raw test score $v$ using the starting-solution score $w$ and target score $u$, both established on the same test set:
\begin{equation}
b(v)=\frac{v-w}{u-w}.
\label{eq:band}
\end{equation}
This maps the starting solution to 0 and the target to 1, with the reproduced baseline providing a separate comparison point on this scale.

\textbf{Submission validity.}
Alongside performance scoring, automated checks assess whether submitted solutions satisfy the required file structure, interfaces, correctness conditions, and resource limits. A model judge agent, distinct from the reviewer agent used during environment construction, also examines the submitted code for violations of task-specific rules. It assigns zero reward to a violating submission regardless of its measured performance.

\textbf{Preventing reward hacking.}
Agents may exploit access to evaluation data or tests to increase their scores without improving their solutions \citep{zhong2025impossiblebench,prasad2026baitbench}. To mitigate such risk, \forge{} restricts access to held-out grading materials. Before submitted code runs, the evaluator loads the test data into memory and removes their on-disk copy. The solver can use development evaluations while working, but held-out scores are not returned during solving. These restrictions do not eliminate all opportunities for exploitation: some environments left other grading materials accessible. We report the observed access failures and the sensitivity of solver results to those failures in \cref{sec:models,app:behaviourfull}.

\Needspace{7\baselineskip}
\section{Building the Corpus}\label{sec:scale}

\subsection{Sourcing}\label{sec:sourcing}

We collect recent preprints separately by domain, then deduplicate the resulting list to broaden disciplinary coverage. Each paper is read in full against six eligibility criteria: a quantitative objective with a stated direction of improvement, a measured baseline, a metric that can be computed by a deterministic script, a method that fits the available compute budget, measurable room for improvement, and a licence permitting redistribution. All source papers are from \nSourceYear{}. These criteria determine which papers enter construction; they do not guarantee that an environment can be built. \Cref{app:screen} gives additional selection and acceptance details, and \cref{sec:funnel} reports the counts.

\subsection{Construction Funnel}\label{sec:funnel}

The main construction run used \opusfive{} for the author agent, reproducer agent, and reviewer agent, with separate sessions and role-specific inputs. \Cref{tab:funnel} traces the source papers through eligibility and construction. We stopped after \nDelivered{} accepted environments; only \nReachDesign{} of the \nCandidates{} eligible papers entered this run, so the counts do not estimate the yield from all eligible papers.

\begin{table}[t]
\centering
\small
\caption{
Main-run construction funnel. Counts include environments
that passed after repairs. The additional \nAblSolN{}
environments built with \gptfivesol{} are outside this funnel.
}
\label{tab:funnel}
\begin{tabular}{lc}
\toprule
Stage & Count \\
\midrule
Source papers screened & \nScreened{} \\
Papers eligible for construction & \nCandidates{} \\
Construction attempts begun & \nReachDesign{} \\
\midrule
Passed Design & \nPassDesign{} \\
Passed Build & \nPassBuild{} \\
Passed Independent Reproduction & \nPassReproduce{} \\
Passed End-to-End Testing & \nPassSmoke{} \\
\midrule
Included in the main corpus & \nDelivered{} \\
\bottomrule
\end{tabular}
\end{table}

\begin{table}[t]
\centering
\small
\caption{Examples of environments from the main construction run.}
\label{tab:corpus_examples}
\setlength{\tabcolsep}{4pt}
\renewcommand{\arraystretch}{1.1}
\resizebox{0.95\textwidth}{!}{
\begin{tabularx}{0.99\textwidth}{@{}>{\raggedright\arraybackslash}p{4.1cm}Y>{\raggedright\arraybackslash}p{5.5cm}@{}}
\toprule
Field & Task & Evaluation objective \\
\midrule
Robotics & Choose footholds on a terrain map & Increase distance in six steps \\
\addlinespace[3pt]
Life Sciences & Propose molecules for a binding pocket & Improve docking scores \\
\addlinespace[3pt]
Earth, Climate \& Space & Estimate weather from partial observations & Reduce state-estimation error \\
\addlinespace[3pt]
Statistics \& Inference & Estimate effects without a control series & Reduce treatment-effect error \\
\addlinespace[3pt]
Software Engineering \& Formal Methods & Generate tests within a fixed budget & Trigger more distinct failures \\
\addlinespace[3pt]
Computational Physics & Infer a plasma boundary from magnetic data & Reduce boundary-estimation error \\
\bottomrule
\end{tabularx}}
\end{table}

Among accepted environments, \nRepairAnyPct\% required at least one revision after failing a construction check. The funnel therefore reports passage after any permitted repairs.
Analysis of repair records identified recurring defects in task instructions and evaluation, summarized in \cref{tab:repair_summary}. Construction used a median of 170 million model tokens per accepted environment. Across construction, 99.6\% of input tokens were cache reads.
\Cref{app:repair} gives the repair breakdown and construction costs.

\begin{table}[t]
\centering
\small
\caption{Most common construction defects identified by model-based coding of repair passages. Percentages are shares of repair passages; categories overlap.}
\label{tab:repair_summary}
\resizebox{0.7\textwidth}{!}{
\begin{tabular}{lr}
\toprule
Issue identified & Share of repair passages \\
\midrule
Checks unable to detect the intended failure & 66\% \\
Incorrect or incomplete instructions & 63\% \\
Checks inspecting a different copy of the code & 54\% \\
Score gains without performing the intended work & 37\% \\
\bottomrule
\end{tabular}}
\end{table}

We then replaced the construction model with \gptfivesol{} and built \nAblSolN{} additional environments using the same queue and checks. This yields \nTotalDelivered{} environments overall. The two construction sets are unpaired by source paper and field; the second run tests whether the procedure works with another backend, rather than estimating a paired model effect (\cref{app:author}).

\subsection{Corpus Composition and Contents}
\label{sec:corpusdata}

The full corpus contains \nTotalDelivered{} environments across \nFields{} research fields. The \nDelivered{} environments from the main construction run include \nFieldMin{}--\nFieldMax{} environments per field. Coverage extends beyond machine learning to areas including robotics, computational physics, statistics, and software engineering.
\Cref{tab:corpus_examples} showcases example tasks and evaluation objectives from six research fields.
\Cref{tab:glance} reports the full field distribution.

For each environment, we retain the task instructions, runnable code, code for generating task data, starting solution, reproduced baseline implementation, and evaluation configuration. Environments provide a development set for experimentation and a held-out test set for scoring. Construction records document feedback from the reviewer agent and repairs.

The corpus is accompanied by quality assessments and records of agent experiments. These include human and automated ratings with written justifications, agent transcripts, saved programs, experiment logs, and held-out scores. Together, they support inspection of both the constructed environments and the sequence of changes agents make while solving them.

\paragraph{Example environment.}
In one weather-estimation environment, the solver estimates a state from a small ensemble and noisy observations of half its components. The starting solution uses an ensemble Kalman method using raw sample covariance. The solver can test changes on 160 development instances; grading uses 324 held-out instances from different simulation conditions. The objective is to minimize mean estimation error. Held-out test errors are 1.786 for the starting solution and 1.390 for the independently reproduced paper method. With a target score of 1.156, normalization maps the starting solution to 0 and the reproduced baseline to approximately 0.629 (\cref{eq:band}).

\subsection{Quality}\label{sec:review}\label{sec:human}

\begin{figure}[t]
\centering
\includegraphics[width=0.67\textwidth]{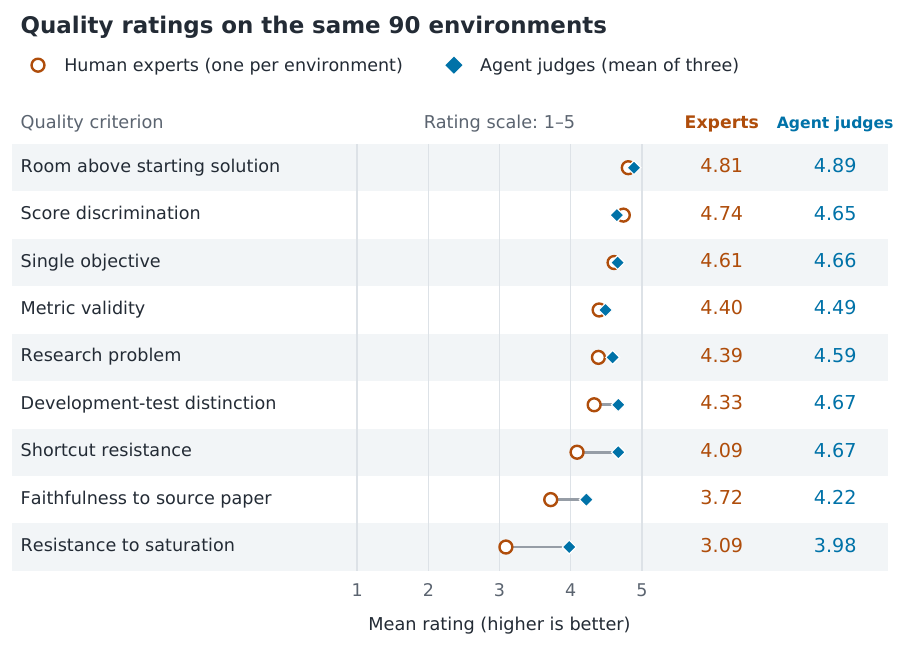}
\caption{Quality ratings from human experts and agent judges on the same 90 environments, with one expert and three judges per environment. Points show means on a 1--5 scale (higher is better). Agent ratings are averaged within environments; rows are ordered by expert mean. Only experts received available solver results. \Cref{app:rubric,tab:human} give the rubric and per-dimension standard deviations.}
\label{fig:expert}
\end{figure}

Three agent judges rated all \nDelivered{} main-run environments, and human domain experts rated \nHumanEnvs{}.
Both groups used the same nine-dimension rubric, with each dimension rated from 1 to 5.
On the expert-reviewed subset,
the means across environments and rubric dimensions were \nHumanMean{} and \nPanelMeanThree{} for human experts and the three-judge panel, respectively.
Experts rated room above the starting solution (\nHheadroom{}) and discrimination between attempts (\nHdiscriminating{}) highest. These ratings support the environments' core measurement properties (\cref{fig:expert,tab:human}).

Experts were more critical of saturation, shortcut resistance, and fidelity to the paper. The widest difference concerned whether one idea could exhaust a task: experts assigned \nHnotsaturable{}, compared with \nPnotsaturable{} from the agent panel. This comparison also reflects unequal information: some experts cited the solver-result summaries withheld from agent judges. Fidelity disagreements often concerned simplifications made to fit the compute budget. Expert-review concerns are summarized in \cref{app:gaps}.

Removing the most generous judge lowers the panel mean to \nPanelMeanTwo{} without eliminating the largest disagreements. For the additional \nAblSolN{} environments built with \gptfivesol{}, all three judges give mean ratings within a fifth of a point of their main-run ratings (\cref{app:author}).
Under the same rubric,
agent judges rated our environments higher than manually constructed RSI-Exam tasks on distinguishing solution quality and resisting shortcuts, whereas RSI-Exam received higher ratings on resistance to saturation (\cref{tab:handbuilt}).
Review ratings are not used to select or weight solver outcomes.

\section{Experiments}\label{sec:found}\label{sec:models}

We use the environments to measure improvement over successive attempts and relative to the reproduced baselines, and analyze how agents explore alternatives, assess changes, and select solutions.

\subsection{Experimental Setup}\label{sec:solversetup}

Before inspecting scores, we randomly sampled \nCommonEnvs{} environments from the main construction run, covering all \nFields{} fields. Four solver agents use the same Claude Code CLI harness, with only the model endpoint changed: \opusfive{}, \gptfive{}, \deepseek{}, and \haikufive{}.

Each model runs \nRoundsN{} successive sessions per environment, yielding \nPanelChains{} model--environment chains and \nPanelGens{} sessions per model.
Sessions begin with fresh contexts. The first receives the starting solution; subsequent sessions inherit the previous session's submitted solution, saved versions, experiment log, and notes. Sessions end when the agent stops or at \nRoundBudgetMin{} minutes.
Almost all ended when the agent stopped. This budget is separate from each environment's budget for running a candidate method. Models exchange no records in the main experiment, and their weights remain fixed.

We report held-out scores of submitted solutions after each session, using the final submission after session~\nRoundsN{} as the primary outcome. We also replay every saved intermediate version on the held-out test set to measure progress within sessions and assess whether agents submit their best saved solutions.
Comparisons use normalized scores and within-environment ranks; bootstrap intervals resample environments. \Cref{app:setup} gives further protocol details, and \cref{tab:effort} reports saved versions and output tokens.

\begin{table}[t]
\centering
\small
\caption{Submitted-solution scores on 120 environments, normalized to the starting solution (0) and target (1). Session~3 confidence intervals use bootstrap resampling of environments. Mean rank averages within-environment ranks at session~3 (1 is best).}
\label{tab:outcome}
\setlength{\tabcolsep}{4pt}
\begin{tabular}{lccccc}
\toprule
& \multicolumn{3}{c}{Median normalized score $\uparrow$} & \multicolumn{2}{c}{Session 3} \\
\cmidrule(lr){2-4}\cmidrule(lr){5-6}
Model & Session 1 & Session 2 & Session 3 & 95\% CI & Mean rank $\downarrow$ \\
\midrule
\opusfive{}  & 0.754 & 0.791 & 0.800 & [0.753, 0.867] & 1.67 \\
\gptfive{}   & 0.673 & 0.747 & 0.761 & [0.723, 0.824] & 1.73 \\
\deepseek{}  & 0.550 & 0.594 & 0.647 & [0.586, 0.711] & 2.97 \\
\haikufive{} & 0.129 & 0.277 & 0.332 & [0.203, 0.451] & 3.63 \\
\bottomrule
\end{tabular}
\end{table}

\begin{figure}[t]
\centering
\includegraphics[width=0.85\textwidth]{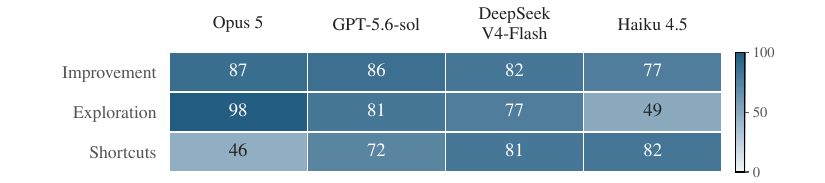}
\caption{Transcript-coded behavior across 360 sessions per model (120 environments, three sessions each). Cell values and color intensity show percentages of sessions assigned to each group by model-based open coding. Groups overlap; ``improvement'' covers developing and checking solutions, without requiring score gains.}
\label{fig:rounds}\label{fig:paradigm}
\end{figure}

\subsection{Performance and Continued Improvement}
\label{sec:panelout}

\Cref{tab:outcome} reports submitted scores after each session. The two strongest models have overlapping median-score intervals; both outperform \deepseek{}, which outperforms \haikufive{}. Within-environment ranks give the same ordering, and paired bootstrap comparisons separate five of the six model pairs. These results describe performance on this corpus under the common harness.
The same solver ordering holds on the \nAblSolN{} environments constructed with \gptfivesol{}, although the two construction sets are unpaired by source paper and field (\cref{tab:ablpanel}).

Most score improvement occurs in the first session for the three stronger models, with smaller gains in later sessions.
With a 0.01 threshold on gains in normalized score, at least one model improves after the first session in \nLaterEnvPct\% of environments. In \nBeatEnvs{} of the \nCommonEnvs{} environments, at least one model surpasses the reproduced baseline in one of its \nRoundsN{} sessions.
These comparisons establish progress within the constructed tasks and against their reproduced baselines (\cref{app:beatref}).

\paragraph{Final solution selection.}
Some final submissions score below an earlier saved version from the same agent, even when the better version passes the automated submission checks. Replaying intermediate versions exposes this difference between discovering a solution and selecting it.

\paragraph{Access to held-out material.}
The transcript coding records access to held-out material or another run's graded record in \nFlaggedChains{} of \nPanelChains{} chains.
We retain these chains in the main results. Excluding them changes each model's median normalized score by at most \nCleanShift{} and preserves the ordering (\cref{app:behaviourfull}).

\subsection{A Three-Session Example}\label{sec:chainexample}

\begin{figure}[htbp]\centering
\includegraphics[width=0.87\textwidth]{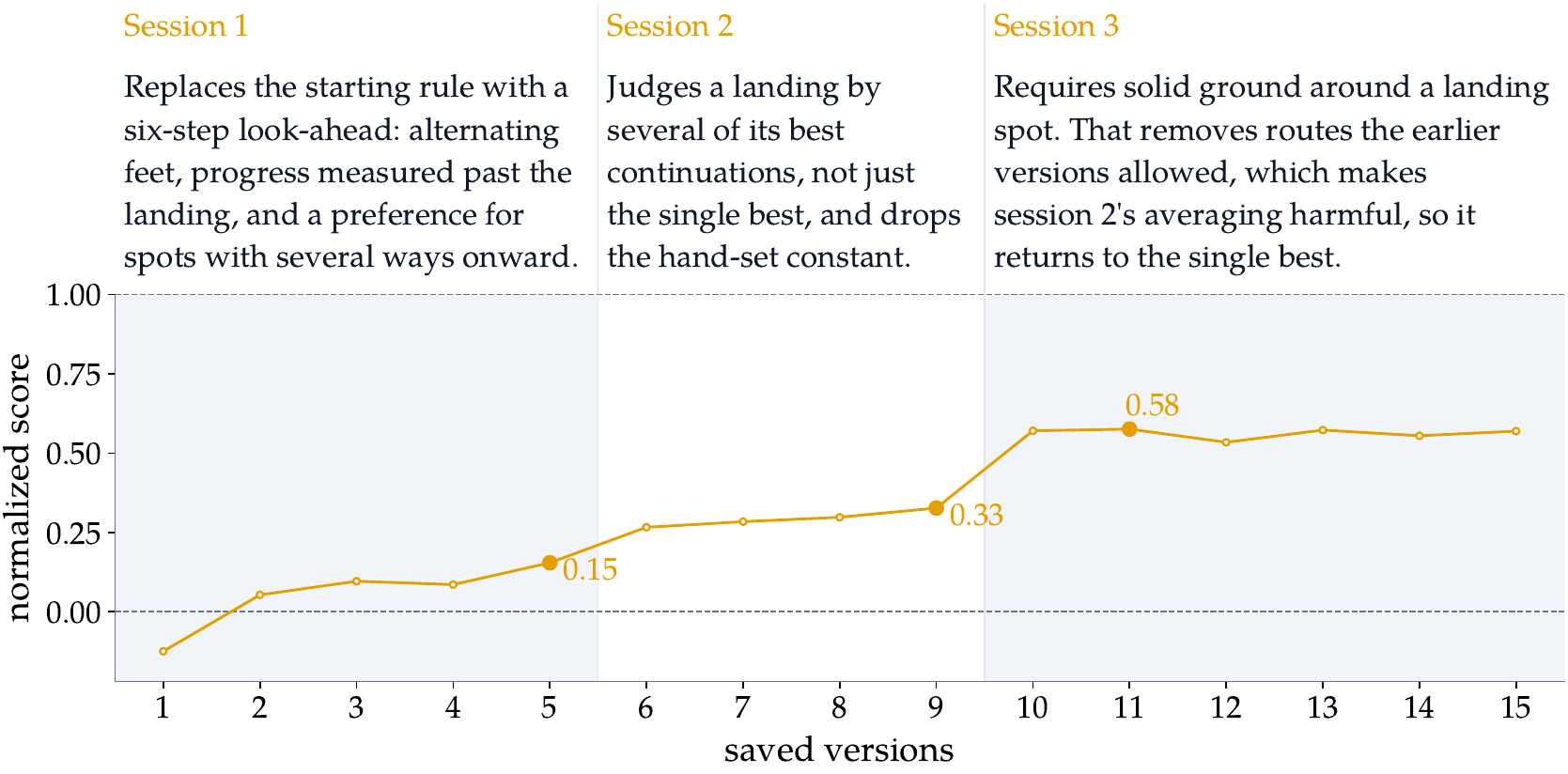}
\caption{A three-session \opusfive{} chain on a foothold-selection task. Open points are saved solutions graded on the held-out test set; filled points mark final submissions, with normalized scores of 0.15, 0.33, and 0.58. The annotations summarize the changes recorded in each session. The third session adds a ground-support check and reverses the second session's averaging rule.}
\label{fig:chain}
\end{figure}

The aggregate scores in \cref{tab:outcome} summarize performance across \nCommonEnvs{} environments. \Cref{fig:chain} shows one \opusfive{} chain in detail: every saved solution graded over three successive sessions and the changes described in the agent's records.

In this environment, a humanoid crossing stepping stones, pillars, stairs, and gaps must choose its next foothold. A submitted solution receives a local elevation map, a six-component state, and a list of admissible locations, and returns the index of one location. The evaluator measures the distance reachable from that foothold within six steps, using held-out terrain generators.

The submitted normalized score rises from 0.15 in session~1 to 0.33 in session~2 and 0.58 in session~3. The second session evaluates each foothold by several of its best continuations, replacing the first session's single-best-continuation rule. In the third session, the agent adds a check that a foothold has sufficient surrounding ground support, removing invalid continuations. With these removed, it returns to the single-best-continuation rule. The trajectory illustrates how an agent can improve by revisiting an inherited design choice.

\subsection{Transcript Analysis}\label{sec:panelbeh}

We analyze the records to characterize how agents work. A model-based coding procedure groups practices into \emph{improvement} (practices for developing and checking solutions), \emph{exploration} (tests of whether a change is useful), and \emph{shortcuts} (score-directed behavior without sufficient evidence of underlying improvement).
For the analysis shown in \cref{fig:paradigm}, \opusfive{} identifies recurring practices without a predefined category list. Each positive assignment requires supporting quotations from the transcript.
Humans did not label these trajectories (\cref{app:behaviourall,app:behaviourfull}).

\Cref{fig:paradigm} summarizes the prevalence of the three behavior groups across models.  All four models frequently modify their solutions, but their evidence-gathering practices differ. Exploration is more common for \opusfive{} than for \haikufive{}, while fitting to the development score follows the reverse pattern. The middle models exchange positions on individual practices.

We check these patterns using a separate analysis with a fixed codebook of ten practices. Three additional model coders apply this codebook to a stratified sample of \nCoderPerModel{} sessions per solver model, obtaining agreement of $\kappa=\nCoderKappa{}$ with the original coder and recovering the same contrast in exploration and shortcuts between \opusfive{} and \haikufive{} (\cref{app:behaviourfull}). A further check uses numeric logs from \nErrBarGens{} sessions with recoverable scores and standard errors. Gains smaller than the reported standard error account for \nErrBarHaiku\% of recorded accepted improvements for \haikufive{}, compared with \nErrBarOpus\% for \opusfive{} (\cref{app:errorbar}).

Among sessions that outperform the reproduced baseline, \nWinStructural\% are coded as
involving work beyond parameter tuning; \nWinTuningOnly\% are tuning-only.
This coding characterizes the work performed in these tasks without establishing novelty to the source paper's field. \Cref{fig:beat} reports mechanism frequencies, and \cref{app:beatref} gives examples of method changes.

\section{Discussion}\label{sec:discussion}

Our results show that published papers can be a practical source of executable research environments across diverse disciplines.
Most accepted environments required repairs before passing the construction checks (\cref{sec:funnel}). Recurring defects included incorrect instructions and checks that could miss failures, showing that automated environment construction involves substantial verification and repair.

The quality reviews clarify what this construction process achieves. Experts rated room above the starting solution and the evaluator's ability to distinguish attempts most highly, supporting the properties needed to measure progress.
Their lower ratings for resistance to saturation and fidelity to the source paper identify a harder challenge: preserving the depth of a research problem when making it executable under a fixed budget.
The disagreement with automated judges suggests that targeted expert review remains valuable for assessing these properties as construction scales.

The solver experiments demonstrate both improvement over successive attempts and gains beyond reproduced baselines. Gains are nevertheless concentrated in the first session for the stronger models. Diminishing gains do not by themselves establish task saturation; they may also reflect limits in how agents search for alternatives or use inherited records.

The trajectories also reveal weaknesses in how agents conduct experiments and retain their results. Lower-scoring models explored less and more often accepted gains smaller than the reported standard error. Across models, some final submissions were worse than previously saved versions, showing that finding a better solution does not ensure it is retained. These observations make experimental judgment and final selection concrete targets for future interventions. The saved programs, measurements, and handover records provide a basis for evaluating such interventions; the present associations do not establish which practices cause higher performance.

\paragraph{Limitations.}\label{sec:limits}
Tasks are adapted to bounded compute, so surpassing a reproduced baseline does not establish an advance on the original research problem. Our sessions measure progress on individual tasks; the study does not isolate the effect of inherited records from extra solving time. No trajectories were used for training; the study does not establish broader gains in research capability. Some environments left held-out grading materials accessible during solver runs. Expert review covers \nHumanEnvs{} of \nDelivered{} main-run environments, with one expert each and no human inter-rater estimate; model-generated trajectory labels lack human validation. We do not ablate construction roles, leaving their individual contributions unresolved. Access restrictions limit independent replication of the full corpus (\cref{app:release}).

\Needspace{12\baselineskip}
\section{Conclusion}
\label{sec:conclusion}

We introduced \forge{}, a multi-agent pipeline that turns published papers into executable RSI environments with automated evaluators and independently reproduced baselines. The corpus contains \nTotalDelivered{} environments across \nFields{} fields. Human expert reviews highlight opportunities to improve starting solutions and the evaluators' ability to distinguish solution quality, while identifying challenges in shortcut resistance and fidelity to the source paper. Solver experiments demonstrate gains over successive attempts and beyond reproduced baselines. Trajectories further reveal differences in how agents explore alternatives, assess uncertain gains, and select solutions. \forge{} provides a scalable approach to expanding the supply of research environments for developing and evaluating self-improving agents.

\bibliographystyle{plainnat}
\bibliography{iclr/ref}

\clearpage
\appendix

\section{Construction Details}\label[appendix]{app:method}\label[appendix]{app:construction}

\subsection{Selection and Acceptance}\label[appendix]{app:screen}\label[appendix]{app:access}\label[appendix]{app:gates}

The six sourcing criteria are given in \cref{sec:sourcing}. A task must provide a single ranking of solutions; additional quantities may impose fixed constraints, but unconstrained trade-offs between objectives do not meet this requirement. Screening for room to improve does not establish that the eventual target score is attainable. The source paper's reported performance guides construction but does not replace measurement of the reproduced method within the adapted environment.

The roles and information restrictions are described in \cref{sec:pipeline}. All three construction roles use \opusfive{} in separate contexts in the main run; this separation does not establish that their errors are independent. Acceptance requires automated checks and reviewer-agent approval at every stage. Development and test sets must differ along a documented task condition, rather than only by a random partition. Independent reproduction uses the same execution path and container as solver solutions and must outperform the starting solution. The author agent has at most \nRepairWorst{} repair attempts per stage before rejection. End-to-end checks cover startup, submission, and grading; they do not establish that all grading materials are inaccessible (\cref{app:behaviourfull}).

\subsection{Repairs and Cost}\label[appendix]{app:repair}\label[appendix]{app:yield}

Among the \nDelivered{} accepted main-run environments, \nRepairAnyPct{}\% required repair: \nRepairDesignPct{}\% during design, \nRepairBuildPct{}\% during build, \nRepairReproPct{}\% during independent reproduction, and \nRepairSmokePct{}\% during end-to-end testing. These stage shares overlap. Main \cref{tab:repair_summary} summarizes model-coded defects in author-agent repair passages; its percentages describe passages, not environments, and categories overlap. Median construction cost is \nCostTotalM{} million model tokens per accepted environment. Median per-environment shares are \nCostAuthorPct{}\%, \nCostReviewerPct{}\%, and \nCostReproducerPct{}\% for author, reviewer, and reproducer, respectively. These separately summarized medians need not sum to 100\%.

\section{Corpus Composition and Quality}\label[appendix]{app:corpus}

\subsection{Field Coverage}\label[appendix]{app:fieldcoverage}

\Cref{tab:glance} reports field counts for the \nDelivered{} main-run environments; \cref{app:author} describes the additional \nAblSolN{} environments constructed with a second backend.

\begin{table}[htbp]\centering\small
\caption{Field composition of the \nDelivered{} main-run environments.}\label{tab:glance}
\begin{tabularx}{\textwidth}{@{}Y r@{\hspace{1.5em}}Y r@{}}
\toprule
Field & Count & Field & Count \\
\midrule
Robotics & 14 & Security \& Cryptography & 10 \\
Life Sciences & 13 & Neuroscience & 9 \\
Earth, Climate \& Space & 13 & Optimisation \& Discrete Algorithms & 9 \\
Statistics \& Inference & 13 & Chemistry \& Materials & 9 \\
Networks \& Information Th. & 12 & Graphics \& Rendering & 8 \\
Software Eng.\ \& Formal Methods & 12 & Quantitative Finance & 7 \\
Signal, Imaging \& Control & 11 & Computational Physics & 7 \\
Computer Systems & 10 & Mechanical \& Thermal & 7 \\
Numerical \& Sci.\ Computing & 10 & Machine Learning & 6 \\
\bottomrule
\end{tabularx}
\end{table}

\subsection{Review Protocol and Results}\label[appendix]{app:review}

Three agent judges rated all \nDelivered{} main-run environments: \opusfive{} under Claude Code, \gptfive{} under Codex at high reasoning effort, and \geminiflash{} under Antigravity. Each model--harness pair is one judge. Judges read task instructions and evaluators and can run code, with network access for field knowledge; the author agent's construction transcript, held-out test data, and solver-result summaries are withheld.

Domain experts independent of environment construction, each holding a doctorate in the relevant field, reviewed \nHumanEnvs{} environments, with one expert per environment and a written justification for every rating. All reviews enter the analysis; there is no human inter-rater estimate. Both groups use the same \nJudgeDims{}-dimension, 1--5 rubric (\cref{tab:reviewaxes}) and have network access. Experts receive the source paper after submitting ratings for the eight environment-focused dimensions, then rate faithfulness last. Only experts receive available summaries of solver scores over successive sessions. Neither group receives construction-stage verdicts or reviewer-agent notes. Expert--agent comparisons therefore reflect differences in evidence as well as judgment.

\textbf{Harness sensitivity.} For \geminiflash{}, \nHarnessPairs{} paired dimension ratings on \nHarnessEnvs{} environments have a mean absolute difference of \nHarnessDiff{} points between Claude Code and Antigravity; \nHarnessWithin\% differ by at most one point. Mean ratings are \nHarnessMeanCc{} and \nHarnessMeanAgy{}, respectively. This indicates limited sensitivity for this judge under these two harnesses.

\Cref{tab:human} gives matched expert and agent-panel ratings with standard deviations. Removing \geminiflash{}, which gives the highest mean on every dimension, lowers the panel mean from \nPanelMeanThree{} to \nPanelMeanTwo{}, compared with \nHumanMean{} for experts. Resistance to saturation and shortcut resistance remain the two largest positive gaps (+0.69 and +0.42, respectively). Ratings are not used to select or weight solver outcomes.

\begin{table}[htbp]\centering\small
\caption{Human-expert and agent-panel ratings on the same \nHumanEnvs{} environments. Scores are means on a 1--5 scale; SD denotes standard deviation. Agent ratings are averaged within environments; Gap is the panel mean minus the expert mean.}
\label{tab:human}
\begin{tabular}{lrrrrr}
\toprule
Dimension & Experts & SD & Panel & SD & Gap \\
\midrule
Research problem & 4.39 & 0.84 & 4.59 & 0.32 & $+0.20$ \\
Room above starting solution & 4.81 & 0.51 & 4.89 & 0.31 & $+0.08$ \\
Resistance to saturation & 3.09 & 0.96 & 3.98 & 0.84 & $+0.89$ \\
Faithfulness to source paper & 3.72 & 0.87 & 4.22 & 0.51 & $+0.50$ \\
Metric validity & 4.40 & 0.85 & 4.49 & 0.45 & $+0.09$ \\
Single objective & 4.61 & 0.90 & 4.66 & 0.46 & $+0.05$ \\
Shortcut resistance & 4.09 & 0.97 & 4.67 & 0.55 & $+0.58$ \\
Score discrimination & 4.74 & 0.62 & 4.65 & 0.55 & $-0.09$ \\
Development--test distinction & 4.33 & 0.93 & 4.67 & 0.49 & $+0.34$ \\
\midrule
All nine dimensions & 4.24 & 0.99 & 4.54 & 0.58 & $+0.30$ \\
\bottomrule
\end{tabular}
\end{table}

\subsubsection{Expert--Agent Disagreements}\label[appendix]{app:gaps}

The largest agent-panel minus expert gaps concern resistance to saturation ($+\nGapnotsaturable$), shortcut resistance ($+\nGapungameable$), and faithfulness ($+\nGapfaithful$). Expert reviews illustrate these concerns.

\textbf{Resistance to saturation.} One expert gave the lowest rating after a supplied search method covered about 69\% of the declared improvement band in roughly five CPU-minutes. The expert also cited solver results reaching the target score in the first session, which agent judges did not receive. The disagreement concerns both available evidence and interpretation of the rubric: reaching a normalization target does not rule out further raw gains or establish that the research problem is exhausted.

\textbf{Shortcut resistance.} One expert proposed adapting SQL answers through repeated held-out grading feedback. The intended protocol withholds that feedback and removes test files before submitted code runs (\cref{sec:band}); whether this attack is possible depends on runtime access controls, not task wording alone. Separately, another review found supplied parameters and held-out family definitions that would let a solution branch on those definitions instead of estimating the intended quantity. We treat that exposure as an environment defect. Observed solver access failures are reported in \cref{sec:panelout,app:behaviourfull}.

\textbf{Faithfulness.} An expert assigned 2 against the panel's 4.5: the task retained 50-way zero-shot classification and comparison of audio and text semantics, but replaced real THINGS-EEG recordings, pretrained feature pipelines, and the trimodal BraVL VAE with synthetic low-dimensional arrays and a linear-Gaussian brain model. A bounded-compute adaptation can preserve the broad question while omitting substantive parts of the source method.

\begin{table}[htbp]\centering\small
\caption{Ratings from the same three agent judges on \nHandBuiltN{} manually constructed RSI-Exam tasks and the \nHumanEnvs{} RSI-Forge environments in \cref{tab:human}. Gap is RSI-Forge minus RSI-Exam; the \nHandBuiltDims{} dimensions exclude faithfulness because this comparison has no common source-paper reference. The task sets are unpaired, so the differences do not isolate the effect of automated construction.}
\label{tab:handbuilt}
\begin{tabular}{lrrr}
\toprule
Dimension & RSI-Exam & RSI-Forge & Gap \\
\midrule
Score discrimination              & 2.64 & 4.65 & $+2.01$ \\
Development--test distinction            & 3.23 & 4.67 & $+1.44$ \\
Shortcut resistance                  & 3.92 & 4.67 & $+0.75$ \\
Single objective            & 4.64 & 4.66 & $+0.03$ \\
Room above starting solution                    & 4.87 & 4.89 & $+0.02$ \\
Research problem                & 4.61 & 4.59 & $-0.02$ \\
Metric validity                    & 4.57 & 4.49 & $-0.08$ \\
Resistance to saturation               & 4.41 & 3.98 & $-0.43$ \\
\midrule
all \nHandBuiltDims{}       & \nHandBuiltMean{} & \nHandBuiltOurs{} & $\nHandBuiltGap{}$ \\
\bottomrule
\end{tabular}
\end{table}

\subsubsection{Review Rubric}\label[appendix]{app:rubric}

Human experts and agent judges use the same quality-review rubric, separate from construction acceptance checks and submission-validity checks. \Cref{tab:reviewaxes} reproduces the descriptions for scores 1, 3, and 5. Terms such as ``starter,'' ``ceiling,'' and ``split'' are retained from the rubric wording; in the surrounding text we use starting solution, target score, and development/test sets.

\begin{table}[htbp]\centering\footnotesize
\caption{Descriptions of scores 1, 3, and 5 in the environment-quality rubric, reproduced verbatim. The dimensions are grouped here as in the rating summaries; the human review form presents faithfulness to the source paper last.}
\label{tab:reviewaxes}
\begin{tabularx}{\textwidth}{@{}>{\raggedright\arraybackslash}p{2.9cm} Y Y Y@{}}
\toprule
Axis & A score of 1 & A score of 3 & A score of 5 \\
\midrule
\emph{Real problem} & a toy with no research content & a real problem, narrowly posed & you would be pleased to see a student work on this \\
\addlinespace[3pt]
\emph{Headroom} & the starter is already near the ceiling & a competent effort gains something & clearly large, and you can name where it comes from \\
\addlinespace[3pt]
\emph{Not saturable} & one change takes most of the range & the first change is worth it, more remains & improvement requires several independent ideas \\
\addlinespace[3pt]
\emph{Faithful to the paper} & unrelated to what the paper does & related, but the hard part has been removed & this is the paper's problem, honestly posed \\
\addlinespace[3pt]
\emph{Metric validity} & measures something else & a reasonable proxy & this is the quantity the field would report \\
\addlinespace[3pt]
\emph{Single direction} & two objectives silently traded & one quantity, the constraint is loose & one quantity, one direction, the constraint binds \\
\addlinespace[3pt]
\emph{Resistant to shortcuts} & you can name a specific trick and say why it scores & you suspect one exists but cannot name it & you looked and the scorer closes the routes you tried \\
\addlinespace[3pt]
\emph{Discriminating} & nothing states it and the metric looks noisy & stated, or inferable, but not justified & stated with the room to improve, and the room is the larger of the two \\
\addlinespace[3pt]
\emph{Split is a real axis} & a uniform random partition dressed up as an axis & an axis, mild shift & a real axis; doing well on it means something \\
\addlinespace[3pt]
\bottomrule
\end{tabularx}
\end{table}

\FloatBarrier
\section{Additional Experimental Details}\label[appendix]{app:experiments}

\subsection{Experimental Setup}\label[appendix]{app:setup}

The protocol in \cref{sec:solversetup} uses solver containers with network access restricted to the model endpoint. The saved files described there are the only inheritance between sessions: the live workspace, conversational context, and shell history are not carried over. Development-evaluator calls have no fixed count limit. The final submission is the solution left in the designated method directory when the agent stops or the \nRoundBudgetMin{}-minute session ends; it is inherited regardless of its score. This session limit is separate from each environment's compute budget for running a candidate method.

Every saved version, including abandoned versions, and the final submission are replayed on held-out data. These scores are not returned during solving. A session without a valid measurement remains unscored rather than receiving an imputed score. \Cref{tab:effort} summarizes saved versions and output volume for the main sample.

\paragraph{Starting and target scores.}\label[appendix]{app:anchor}\label[appendix]{sec:beat}
In \cref{eq:band}, the starting-solution score $w$ maps to 0 and the target $u$ to 1; better-than-target scores exceed 1. Construction records identify the data, code, or bound supporting each value and mark provisional values. Targets may use privileged information, an information-theoretic or optimization bound, exact solutions on tractable instances, or a Bayes rate. They need not be attainable, and normalization does not validate them. If the reproduced baseline is too close to the target, construction checks require revision of the target or task conditions.

\begin{table}[htbp]\centering\small
\caption{Activity per session on the \nCommonEnvs{} main evaluation environments. Versions are saved runnable solutions that were graded; output tokens are in thousands. Means include one standard deviation. Token volume is not equivalent to monetary or computational cost across providers.}
\label{tab:effort}\label{tab:cost}
\begin{tabular}{lrrrr}
\toprule
& \multicolumn{2}{c}{Versions graded} & \multicolumn{2}{c}{Output tokens (thousands)} \\
\cmidrule(lr){2-3}\cmidrule(lr){4-5}
Model & Median & Mean & Median & Mean \\
\midrule
\opusfive{}  & 4.0 & $3.6 \pm 1.6$ & 69.3 & $72.2 \pm 22.8$ \\
\gptfive{}   & 2.0 & $2.7 \pm 1.7$ & 22.1 & $22.6 \pm 8.3$ \\
\deepseek{}  & 2.0 & $2.1 \pm 1.1$ & 54.5 & $53.8 \pm 16.3$ \\
\haikufive{} & 2.0 & $2.7 \pm 1.9$ & 30.2 & $29.8 \pm 8.8$ \\
\bottomrule
\end{tabular}
\end{table}

\subsection{Improvements over Reproduced Baselines}\label[appendix]{app:beatref}

At least one session surpasses the reproduced baseline in \nBeatEnvs{} of \nCommonEnvs{} environments. The per-model counts are \nBeatGpt{} for \gptfive{}, \nBeatOpus{} for \opusfive{}, \nBeatDeep{} for \deepseek{}, and \nBeatHaiku{} for \haikufive{}. These counts overlap, and an environment can contribute several successful sessions. The median normalized baseline score is \nRefLost{} where no model surpasses it and \nRefWon{} where at least one does.

\begin{figure}[htbp]\centering
\includegraphics[width=\textwidth]{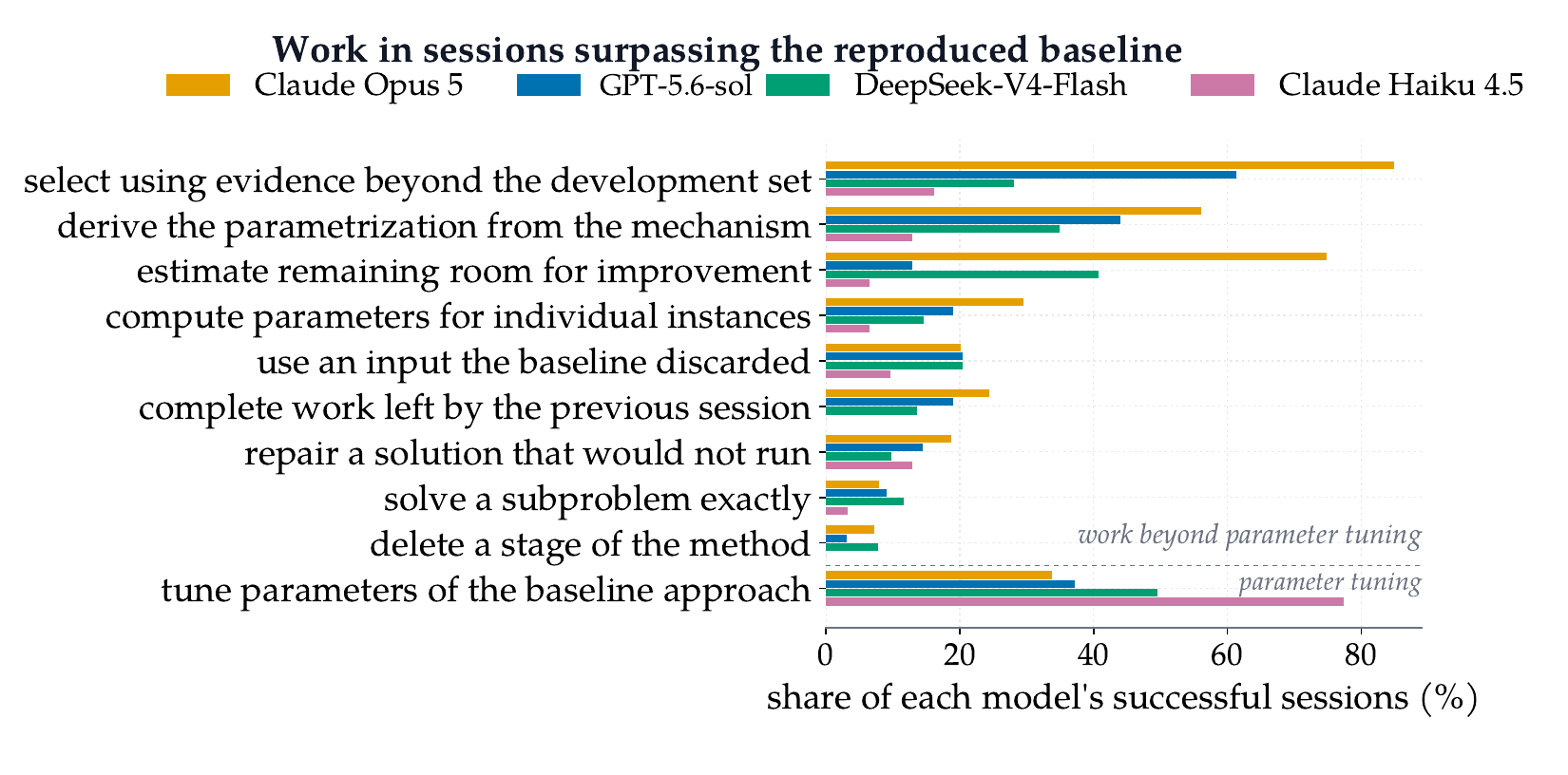}
\caption{Transcript-coded mechanisms as percentages of each model's sessions that surpass the reproduced baseline. The codebook was derived from a sample and fixed before counting. Sessions may receive several labels; nine categories cover work beyond parameter tuning. The labels describe work within the constructed task and do not establish scientific novelty.}
\label{fig:beat}
\end{figure}

The separate mechanism codebook in \cref{fig:beat} applies only to these successful sessions. Across them, \nWinStructural\% receive at least one of the \nWinMechs{} labels beyond parameter tuning, \nWinTuningOnly\% receive only the tuning label, and \nWinNeither\% receive neither. These labels include repairing a non-running solution and evaluating alternatives, so they do not all imply a new method. Tuning can coexist with other work: it appears in \nBeatRetuneOpus\% of successful \opusfive{} sessions and \nBeatRetuneHaiku\% of successful \haikufive{} sessions. The coding does not establish which activity caused a gain or whether the result advances the source paper's field.

Actual method changes include retaining alternating least squares while replacing a non-differentiable tolerance with a smooth surrogate, and integrating over parameters that the reproduced baseline estimated at a single point.

\subsection{Behavior Coding and Analysis}\label[appendix]{app:behaviourall}

Three coding procedures serve different purposes: open coding supplies the behavior groups in \cref{fig:paradigm}; a fixed practice codebook checks those patterns; and the separate mechanism codebook characterizes successful sessions (\cref{app:beatref}). All labels are model-generated, without human validation.

\subsubsection{Coding Procedures}\label[appendix]{app:behaviour}\label[appendix]{sec:xval}

\paragraph{Open coding.} \opusfive{} summarizes each session's changes, evidence, score-directed work, unexpected events, and stopping decision, then groups recurring practices in bounded batches without a predefined category list. Document identifiers are retained through merges, and positive assignments require matching transcript quotations. This yields \nFamMapped{} families mapped to \emph{improvement} (developing and checking solutions, without requiring score gains), \emph{exploration} (testing whether a change is useful), and \emph{shortcuts} (score-directed behavior with insufficient evidence of underlying improvement), plus one ungrouped family for submitting the inherited solution unchanged. A session receives a group label if it has at least one family in that group; labels overlap and need not sum to 100\%.

\paragraph{Fixed practices.} \opusfive{} proposes categories from \nCodeSample{} documents selected with a fixed seed and no supplied category list. The ten-category codebook is then frozen and applied to all available session records (\cref{tab:codebook}). A session can receive multiple labels or none; assignments without a matching quotation are discarded. Quotations make decisions traceable without validating the interpretation or the narrated action. Within-model assignment counts explain little variation in output-token counts, but this does not rule out narration-style differences; additional coders and numeric logs provide complementary checks below.

\begin{figure}[htbp]\centering
\includegraphics[width=\textwidth]{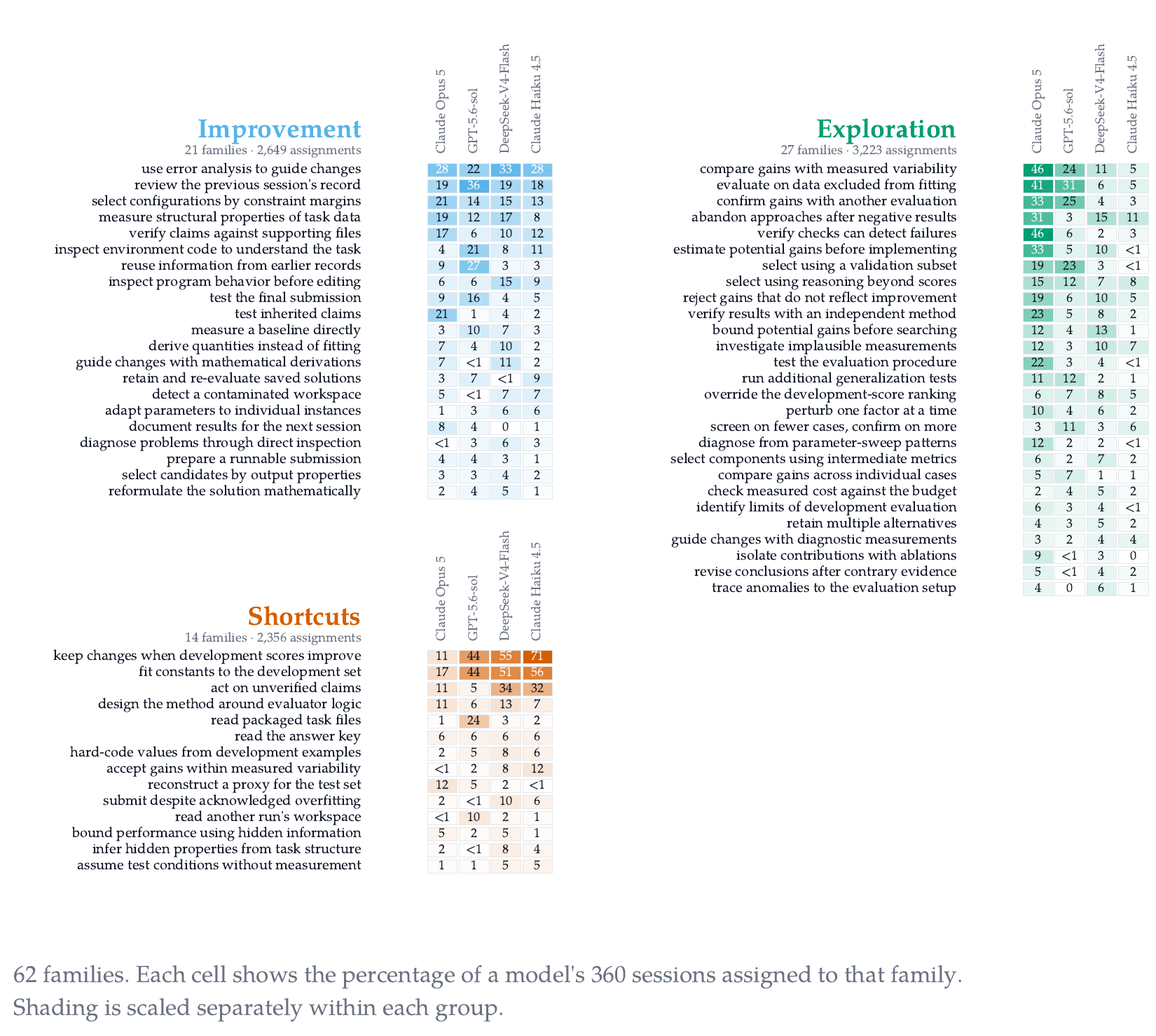}
\caption{Open-coded families grouped as improvement (\nFamImprove{}), exploration (\nFamExplore{}), and shortcuts (\nFamShort{}). Cells show percentages of each model's \nPanelGens{} sessions assigned a family. Labels overlap. Shading is scaled within each group; printed percentages support comparisons across groups. The ungrouped family for submitting the inherited solution unchanged is omitted.}
\label{fig:famheat}
\end{figure}

\begin{table}[htbp]\centering\small
\caption{The ten fixed practices, distinct from the open-coded families in \cref{fig:famheat} and the successful-session mechanisms in \cref{fig:beat}. Labels and operational criteria are paraphrased from the coding scheme; timing requirements and exclusions are retained.}
\label{tab:codebook}\label{app:codebook}\label{app:solvercode}
\setlength{\tabcolsep}{4pt}
\begin{tabularx}{\textwidth}{@{}>{\raggedright\arraybackslash}p{2.2cm}>{\raggedright\arraybackslash}p{3.4cm}Y@{}}
\toprule
Group & Practice & Required evidence \\
\midrule
Improvement & Check inherited results & Before any edit, rerun the inherited solution to reproduce its recorded score or independently derive its central claim. \\
& Use the task specification & Justify the submitted change from a task-document fact or metric property; a parameter-sweep result alone is insufficient. \\
& Improve robustness & Address failures such as degenerate inputs, timing limits, or fallback behavior without targeting the metric. \\
\addlinespace[3pt]
Exploration & Bound potential improvement & Before searching, compute an upper or lower performance bound and decompose the remaining gap to allocate the budget. \\
& Test beyond development data & Evaluate a change beyond the supplied development instances using a stress test or test of invariance. \\
& Compare with error bars & Use an explicit error bar when accepting or rejecting a change; a before-and-after comparison alone is insufficient. \\
& Test controls or independent checks & Check that a measurement detects the claimed effect using a deliberately weakened variant, an incorrect-generator control, or an independent computation. \\
& Reject unreliable gains & Discount a gain because its evidence is circular, depends on a parameter fitted to development data, or changes sign across evaluations. \\
\addlinespace[3pt]
Shortcut & Select by development score alone & Run successive sweeps of individual parameters and treat any favorable development-score change as an improvement. \\
Neither & Environment failure & An environment failure unrelated to the solution method consumes the session. \\
\bottomrule
\end{tabularx}
\end{table}

\subsubsection{Behavior Patterns and Sensitivity Checks}\label[appendix]{app:behaviourfull}

\paragraph{Shortcut labels and access failures.} Common shortcut families include retaining a version because its development score rises and fitting constants to the development set. These can occur within the permitted workflow; the label concerns limited evidence of generalization. Other families cover reading answer keys, grading code, another run's records, or reconstructing tests from descriptions. These are broader than the access-failure criterion: a family may include labels supplied with development data, whereas an access flag requires a record naming held-out material or another run's graded record. The coding cannot exclude unrecorded access.

This narrower criterion flags \nFlaggedChains{} of \nPanelChains{} model--environment chains. They remain in the main results. Excluding flagged chains gives median final normalized scores of \nCleanA{}, \nCleanB{}, \nCleanC{}, and \nCleanD{} for \opusfive{}, \gptfive{}, \deepseek{}, and \haikufive{}, respectively: ordering is unchanged, and no median moves by more than \nCleanShift{}. Filtering is by chain; a flag for one model does not exclude the other models on that environment. In \nWinsRead{} of the \nBeatEnvs{} environments where a model surpasses the baseline, a successful session carries an access flag. This overlap does not attribute the gain to access. Behavior-group percentages use the full sample; only final-score sensitivity uses the exclusion.

\paragraph{Additional model coders.} \sonnetfive{}, \geminiflash{}, and \gptfivesol{} apply the fixed practice codebook to a stratified sample of \nCoderPerModel{} sessions per solver model. Agreement with the original coder is $\kappa=\nCoderKappa{}$ over the \nCoderCats{} categories used by either coder. On these records, the original coder labels exploration in \nCoderExplOpusOne\% of \opusfive{} sessions and \nCoderExplHaikuOne\% of \haikufive{} sessions; additional coders give \nCoderExplOpus\% and \nCoderExplHaiku\%. Shortcut rates reverse the contrast: \nCoderShortOpusOne\% versus \nCoderShortHaikuOne\% originally, and \nCoderShortOpus\% versus \nCoderShortHaiku\% with additional coders. This corroborates the contrast without establishing individual-label correctness. These rates use the fixed codebook on the sampled subset, whereas \cref{fig:paradigm} uses open-coded families on all \nPanelGens{} sessions per model.

\subsubsection{Gains Smaller than the Reported Standard Error}\label[appendix]{app:errorbar}

For \nErrBarGens{} sessions, evaluator logs provide both scores and standard errors. We classify an accepted gain as smaller than the reported error when the selected version improves on the best previous version by less than the standard error attached to the selected score. This describes selection; it is not a significance test for the difference between correlated scores. Accepted improvements in these recoverable sessions form the denominator, not all sessions or saved versions. The rates are \nErrBarHaiku\% for \haikufive{} and \nErrBarOpus\% for \opusfive{}. This agrees with their fixed-codebook ordering, while measuring a different behavior: the codebook concerns selection from development-score changes without sufficient supporting evidence.

\subsection{Environments by Backend Model}\label[appendix]{app:author}

Replacing \opusfive{} with \gptfivesol{} in the same sourcing queue, checks, and repair procedure produces \nAblSolN{} additional environments, for \nTotalDelivered{} overall. The comparison uses accepted environments only and does not measure construction yield. Sets are unpaired by source paper and field, so differences can reflect task composition as well as backend choice. All three judges' mean ratings differ by less than 0.2 points (\cref{tab:abljudge}). Solvers retain the same ordering (\cref{tab:ablpanel}); this extends the observation to a second construction set without isolating a causal backend effect.

\begin{table}[htbp]\centering\small
\caption{Agent-judge ratings by construction backend. Cells average over rubric dimensions, then environments; $n$ counts rated environments. For this comparison, \geminiflash{} uses Claude Code; \cref{app:review} compares its ratings under Claude Code and Antigravity.}
\label{tab:abljudge}
\begin{tabular}{lrrrr}
\toprule
& \multicolumn{2}{c}{\opusfive{}} & \multicolumn{2}{c}{\gptfivesol{}} \\
\cmidrule(lr){2-3}\cmidrule(lr){4-5}
Agent judge & Mean & $n$ & Mean & $n$ \\
\midrule
\opusfive{} (Claude Code) & \nAblJudgeOpusCor{} & \nAblJudgeOpusCorN{} & \nAblJudgeOpusSol{} & \nAblJudgeOpusSolN{} \\
\gptfive{} (Codex) & \nAblJudgeGptCor{} & \nAblJudgeGptCorN{} & \nAblJudgeGptSol{} & \nAblJudgeGptSolN{} \\
\geminiflash{} (Claude Code) & \nAblJudgeGemCor{} & \nAblJudgeGemCorN{} & \nAblJudgeGemSol{} & \nAblJudgeGemSolN{} \\
\bottomrule
\end{tabular}
\end{table}

\begin{table}[htbp]\centering\small
\caption{Median final submitted normalized scores after three sessions by construction backend. The main-backend column repeats session~3 of \cref{tab:outcome}; $n$ counts chains with all three sessions scored. Construction sets are unpaired.}
\label{tab:ablpanel}
\begin{tabular}{lrrrr}
\toprule
& \multicolumn{2}{c}{\opusfive{}} & \multicolumn{2}{c}{\gptfivesol{}} \\
\cmidrule(lr){2-3}\cmidrule(lr){4-5}
Solver model & Median & $n$ & Median & $n$ \\
\midrule
\opusfive{}  & 0.800 & 120 & 0.784 & 30 \\
\gptfive{}   & 0.761 & 120 & 0.736 & 30 \\
\deepseek{}  & 0.647 & 120 & 0.564 & 30 \\
\haikufive{} & 0.332 & 120 & 0.363 & 30 \\
\bottomrule
\end{tabular}
\end{table}

\FloatBarrier
\section{Release and Access}\label[appendix]{app:closing}\label[appendix]{app:reporting}\label[appendix]{app:release}

The corpus is available on request. Access is gated because the sealed test split and grading key make grading possible and also make scores obtainable without solving the task. We do not release the sourcing procedure, per-paper verdicts, or rejection reasons, to avoid attaching our assessments of benchmark suitability to individual preprints and their authors.

\end{document}